\documentclass{article}

\usepackage{PRIMEarxiv}
\usepackage[utf8]{inputenc} 
\usepackage[T1]{fontenc}    
\usepackage{hyperref}       
\usepackage{url}            
\usepackage{booktabs}       
\usepackage{amsfonts}       
\usepackage{nicefrac}       
\usepackage{microtype}      
\usepackage{lipsum}
\usepackage{fancyhdr}       
\usepackage{graphicx}       
\graphicspath{{media/}}     
\usepackage{orcidlink}
\usepackage[round]{natbib}
\usepackage{graphicx}
\usepackage{float}

\title{Tamil-Llama: A New Tamil Language Model Based on Llama 2
}

\author{
  Abhinand Balachandran \orcidlink{0009-0004-9692-8432} \\
  \texttt{abhinandb.ml@gmail.com} \\
}

\begin{document}
\maketitle

\begin{abstract}
Language modeling has witnessed remarkable advancements in recent years, with Large Language Models (LLMs) like ChatGPT setting unparalleled benchmarks in human-like text generation. However, a prevailing limitation is the underrepresentation of languages like Tamil in these cutting-edge models, leading to suboptimal performance in diverse linguistic contexts. This paper addresses this lacuna, enhancing the open-source LLaMA model with an addition of 16,000 Tamil tokens, aiming to achieve superior text generation and comprehension in the Tamil language. We strategically employ the LoRA methodology for efficient model training on a comprehensive Tamil corpus, ensuring computational feasibility and model robustness. Moreover, we introduce a Tamil-translated version of the Alpaca dataset and a subset of the OpenOrca dataset tailored for instruction fine-tuning. Our results showcase significant performance improvements in Tamil text generation, with potential implications for the broader landscape of LLMs in Indian languages. We further underscore our commitment to open research by making our models, datasets, and code\footnote{GitHub Repository: \href{https://github.com/abhinand5/tamil-llama}{https://github.com/abhinand5/tamil-llama}} publicly accessible, fostering further innovations in language modeling.
\end{abstract}

\section{Introduction}

The past few years have been transformative for language modeling, with groundbreaking advances and monumental achievements. At the forefront of this revolution was OpenAI's ChatGPT \citep{chatgpt}, which not only raised the bar in language modeling performance but also underscored the immense societal implications of such technologies. Alongside ChatGPT, various Large Language Models (LLMs) have consistently demonstrated exceptional prowess in natural language understanding and generation, heralding a new era in computational linguistics.

Central to the functionality of these modern LLMs is the Transformer architecture, a cornerstone concept brought to the limelight by "Attention is All You Need" \citep{vaswani2017attention}. This innovation transformed our approach to sequence-based tasks, catalyzing pivotal models like BERT \citep{devlin2019bert} and redefining best practices in Natural Language Processing (NLP).

Subsequent developments, particularly the Generative Pre-trained Transformer (GPT) \citep{radford2018improving}, showcased the profound potential of unsupervised pre-training on vast datasets. Models like GPT-3 and its successor, GPT-4 \citep{openai2023gpt4}, have redefined benchmarks and fueled a renaissance in natural language understanding and generation. Beyond their technical prowess, they have prompted a renewed vigor in exploring the limits of Artificial General Intelligence (AGI). These advancements, paired with exemplary performance in numerous applications, have galvanized the NLP community, sparking widespread application and research from sentiment analysis to machine translation.

However, progress is not without its pitfalls. The elite LLMs, despite their remarkable capabilities, grapple with challenges—primarily, their proprietary nature, which constricts open research. Furthermore, an English-centric bias and the enormous computational requirements for training such behemoths further accentuate the call for more accessible and diverse solutions.

In response, the open-source community has championed the creation of models like LLaMA \citep{touvron2023llama} and Mistral \citep{jiang2023mistral}. Such models, despite their compact nature, challenge the hegemony of giants like ChatGPT in select benchmarks, heralding a promising direction for future research.

However, as robust as these models, like LLaMA and Mistral, might be, their proficiency in generating coherent text in Tamil and several other Indian languages remains noticeably deficient. A fundamental limitation lies in their minimal vocabulary of Tamil characters, which is essential for effective text encoding and generation. This paper aims to bridge this gap by augmenting the existing LLaMA models' vocabulary with an additional 16,000 Tamil tokens, markedly enhancing their capability in processing and producing Tamil content. This method draws inspiration from a parallel endeavor in the Chinese adaptation of LLaMA, as documented in \cite{cui2023efficient}. To ensure efficient pre-training and fine-tuning while maintaining computational feasibility, we leverage the LoRA \citep{hu2021lora} methodology. We aspire that this initiative catalyzes further research endeavors, refining LLaMA and other open-source models tailored for Indian languages. A succinct overview of the principal contributions of this paper is as follows:

\begin{itemize}
\item We bolster the LLaMA model's encoding and decoding proficiencies for Tamil by incorporating an additional 16,000 Tamil tokens, thereby expanding its vocabulary.
\item Through the LoRA methodology, the augmented model undergoes training on an extensive Tamil corpus, resulting in a marked enhancement of its text generation capabilities relative to its predecessor models.
\item We present a Tamil-translated version of the original Alpaca dataset \citep{alpaca}, paired with a subset of the OpenOrca \citep{OpenOrca} dataset, both curated for instruction fine-tuning in Tamil.
\item Our newly trained instruction and chat models, built upon the Alpaca and OpenOrca datasets, demonstrate notable advancements in performance for the Tamil language compared to other open-source language models.
\item To stimulate continuous innovation and broader adaptability, we grant public access to the models, datasets, and associated code, inviting further exploration and encouraging the refinement of LLaMA models for diverse languages.
\end{itemize}

\section{Related Work}

Within the broad field of Natural Language Processing (NLP), the advent of Large Language Models (LLMs) marks a transformative moment. These models have heralded new capabilities in understanding, generating, and processing various human languages, underpinning innovations from automated content creation to nuanced sentiment analysis. While their proficiency in mainstream languages like English is widely recognized and leveraged, a disparity exists in their performance and availability for numerous non-European languages.

Tamil, a language with ancient roots and spoken by a substantial global population, epitomizes this disparity. Despite its linguistic depth and cultural significance, dedicated pre-trained LLMs for Tamil are conspicuously underrepresented. Most current offerings are generic, multipurpose LLMs, which do not cater specifically to the unique attributes of the Tamil language.

A survey of the existing literature reveals that many attempts to cater to the Tamil language through LLMs rely heavily on multilingual models. Works such as \cite{scao2022bloom}, \cite{mGPT}, and \cite{lin2022fewshot} have all ventured into this domain. However, it is crucial to note that, except "GPT-2 Tamil" by \cite{abinayam/gpt-2-tamil}, all these models are not exclusive to Tamil. While they can process Tamil to a certain extent, their capabilities are inherently limited. This limitation arises because the training data for these models often comprise a low fraction of Tamil content relative to other languages. Consequently, the nuances and intricacies specific to Tamil are often lost, leading to suboptimal performance.

The effort by \cite{abinayam/gpt-2-tamil} represents a notable deviation from this trend. Here, the GPT-2 base model, equipped with 117 million parameters as outlined in \cite{radford2019language}, was fine-tuned with a focus on Tamil, using both the Oscar dataset \citep{caswell2020language} and The IndicNLP \citep{kunchukuttan2020indicnlp} dataset. This approach signifies a targeted attempt to adapt LLM capabilities for the Tamil language specifically.

However, the broader landscape of Tamil-specific LLM research remains relatively uncharted. This context underscores the motivation for our present research. We endeavor to delve deeper into this space, addressing existing shortcomings and advancing the capabilities of LLMs tailored for Tamil.

\section{Tamil LLaMA} 

\subsection{Datasets Used}

The development of Tamil-LLaMA involved using several different datasets, each chosen for specific parts of the training and fine-tuning process. This approach was vital to ensure the model's effectiveness across various tasks.

\subsubsection{Datasets used for Pre-Training}\label{subsec:pretraining-datasets}

For the initial pre-training phase of LLaMA 2 \citep{touvron2023llama}, we mainly used the CulturaX dataset \citep{nguyen2023culturax}. This dataset is a combination of many popular datasets, including the Oscar dataset \citep{caswell2020language}. Out of the 4.72 million documents in CulturaX, we selected 600k documents (12 GB) for training. This choice was made to manage training costs while aiming for high performance. Our approach was successful, as the model showed strong results in text completion tasks even with this smaller dataset.

\subsubsection{Datasets used for Instruction Tuning}\label{subsec:finetuning-datasets}

The "Instruction Tuning" phase was a pivotal stage in refining LLaMA's proficiency in precisely adhering to textual instructions. For this enhancement, we incorporated a translated version of the Stanford Alpaca dataset \citep{alpaca}, comprising 52,000 instructions. Concurrently, we integrated a specialized no-code section from the OpenOrca dataset \citep{OpenOrca}, which consists of around 93,000 instructions. The deliberate focus on no-code instructions was to streamline the training process, eliminating the intricacies presented by coding instructions during translation.

To ensure translation uniformity and accuracy across the datasets, the Google Translation API service was our tool of choice. We meticulously translated the entirety of the Alpaca dataset while also applying a similar methodology to the OpenOrca subset.

We believe that leveraging diverse datasets has bolstered LLaMA's enhanced capability to discern and generate contextually pertinent responses across a spectrum of prompts.

\subsection{Background on the LLaMA Models}

Introduced by \cite{touvron2023llama}, LLaMA has emerged as an essential milestone in the world of open-source large language models (LLMs), with the renowned Transformer architecture \citep{vaswani2017attention} as its foundation. While it draws inspiration from models like GPT for its basic structure—comprising an embedding layer and multiple transformer blocks—LLaMA has its unique features. LLaMA has brought forward several innovative techniques such as pre-normalization \citep{zhang2019root}, SwiGLU activation \citep{shazeer2020glu}, and rotary embeddings \citep{su2022roformer}. Offered in sizes ranging from 7B (7 Billion) to 65B (65 Billion) parameters, LLaMA has been trained on a rich mixture of content sources, including web pages, books, and academic papers. Its strong performance on benchmarks, especially given its relatively compact size compared to other models, has made it a noteworthy contender in the LLM landscape, drawing considerable attention in the AI research community.

Building upon its predecessor's foundation, LLaMA 2 \citep{touvron2023llama2} introduces monumental enhancements to the LLaMA lineage. With a dataset expanded by 40\% relative to LLaMA 1, the models under LLaMA 2 exhibit an enriched comprehension of diverse content, leading to improved text generation. An extended context length of 4,096 tokens empowers LLaMA 2 to process and understand more extensive textual segments, significantly benefiting tasks such as translation and intricate question answering. Another pivotal innovation in LLaMA 2 is adopting the grouped-query attention mechanism \citep{ainslie2023gqa}, facilitating faster inference despite its expanded size compared to LLaMA 1.

In the course of our research, we made a conscious choice to employ LLaMA 2 as our primary language model. Several factors influenced this decision. Firstly, LLaMA 2 is a recent addition to the lineage of Large Language Models, which implies that it benefits from the latest advancements in model training and architectural innovations. This recent launch incorporates the most up-to-date techniques and methodologies. Secondly, compared with its predecessor, LLaMA 1, the enhancements in LLaMA 2 are undeniably compelling. These improvements are not just incremental; they represent substantial strides in areas such as data exposure, context length, and attention mechanisms. The evolution from LLaMA 1 to LLaMA 2 is emblematic of the rapid advancements in the field, and by leveraging the latter, we aimed to ensure our research was grounded in the most cutting-edge tools available.

\subsection{Expansion of Tamil Vocabulary}

LLaMA 2, as outlined in the seminal work of \cite{touvron2023llama2}, is backed by an expansive pre-training corpus of 2 Trillion tokens. A detailed linguistic analysis of this vast corpus reveals a striking imbalance in language representation. An overwhelming 89.7\% of the tokens are sourced from English, with other European languages collectively contributing to nearly 10\% of the dataset. In stark contrast, diverse languages such as Tamil and Hindi represent a meager presence, with their combined token count along with other under-represented languages accounting for less than 0.21\%.

This skewed distribution raises concerns about the genuine multilingual and cross-lingual capabilities of LLaMA 2. While it is evident that the model is proficient in several European languages, its ability to comprehend and generate content in languages like Tamil needs to be improved substantially. Our preliminary experiments further underscored this limitation. When presented with tasks in Tamil, LLaMA 2 exhibited a remarkable lack of coherence in its responses. In fact, its performance was notably inferior to smaller models, underscoring a noticeable shortcoming in LLaMA 2's coverage of worldwide languages. There is a clear need for the open-source community to focus on languages like Tamil, spoken by millions globally across multiple countries.

To bolster the text generation and understanding abilities of LLaMA 2 in Tamil, we advocate extending its pre-training phase with an expansive Tamil corpus, as recommended by \cite{cui2023efficient}. However, this alone is not sufficient. A limitation arises from LLaMA's existing vocabulary, which has a tiny number of Tamil characters. Although LLaMA can bypass this by encoding unknown tokens, this process considerably lengthens the sequences, leading to substantial delays during encoding and decoding. Typically, a single Tamil character is translated into 3-4 byte tokens. Moreover, these byte tokens are not uniquely purposed for Tamil characters but represent UTF-8 tokens from various languages. This dual role complicates the task for transformer encoders and byte-tokens to understand and capture the nuanced semantics of Tamil characters proficiently. 

To overcome these problems and to enhance the text generation capabilities in Tamil, we propose the incorporation of an additional 16,000 Tamil tokens to the pre-existing vocabulary of the LLAMA 2 model. This methodology echoes the strategies employed in developing Chinese LLaMA \citep{cui2023efficient}. The subsequent steps explain the process of vocabulary extension:

\begin{enumerate}
\item Employ SentencePiece \citep{kudo2018sentencepiece} to train a Tamil Tokenizer on an extensive corpus of contemporary Tamil text, capturing the essence of modern linguistic nuances necessary for coherent communication.
\item Integrate the original tokenizer of the LLaMA 2 model with the vocabulary derived from the newly trained SentencePiece tokenizer. This amalgamation culminates in an augmented tokenizer encompassing an additional 16,000 Tamil tokens, leading to an aggregated vocabulary size of 48,000 (32,000 original + 16,000 new).
\item Drawing parallels from \cite{cui2023efficient}, the LLaMA model is then tailored to accommodate the Tamil LLaMA tokenizer. This modification necessitates resizing the word embeddings and the language model head from a matrix shape V × H to V' × H. Herein, V represents the original vocabulary size of 32,000, whereas V' signifies the extended size of 48,000. Importantly, this adjustment ensures the preservation of the embeddings associated with the original vocabulary by appending the new rows to the concluding segments of the initial embedding matrices.
\end{enumerate}

\begin{figure}[h]
    \centering
    \caption{Tokenizer comparisons between original LLaMA and Tamil LLaMA.}
    \includegraphics[width=0.9\linewidth]{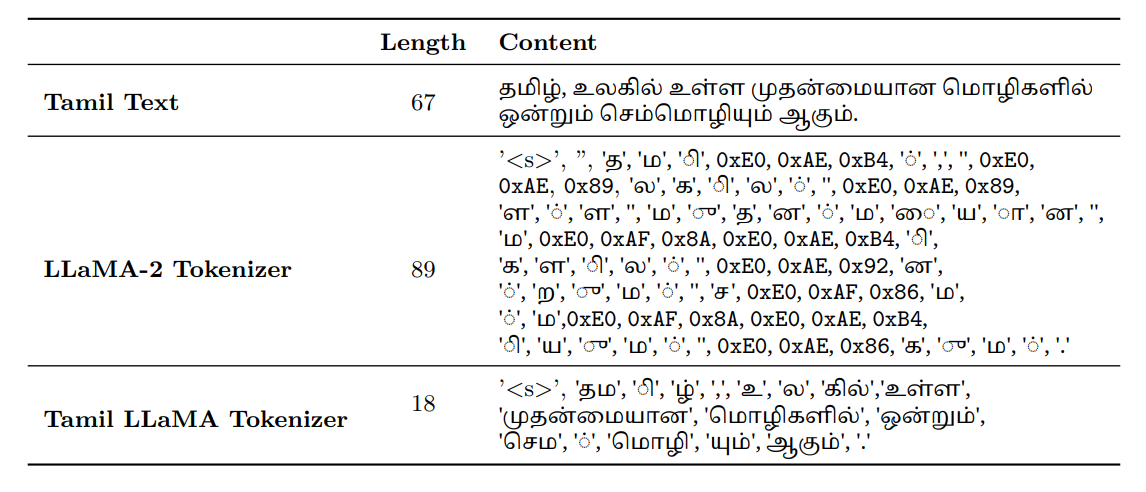}
    \label{fig:tokenizer-comparison}
\end{figure}

In Figure \ref{fig:tokenizer-comparison}, we can see that the Tamil LLaMA tokenizer needs only 20\% to 25\% of the tokens that the original LLaMA model uses to encode Tamil text. This makes the Tamil LLaMA much more efficient. With this crucial update, the model can handle over three times more information and works three times faster. In conclusion, our modifications to LLaMA 2 significantly bolster its capabilities in understanding and generating Tamil content. By adding 16,000 Tamil tokens, we ensure a more efficient and nuanced representation. The new Tamil LLaMA tokenizer drastically reduces the required tokens, making encoding more efficient.

\subsection{Pre-Training Phase}

In order to harness the full potential of the expanded vocabulary of Tamil LLaMA, a robust pre-training phase is implemented using a comprehensive Tamil text corpus. The datasets utilized during this training phase are detailed in \ref{subsec:pretraining-datasets}.

\paragraph{Causal Language Modelling Approach}
The central mechanism for this pre-training is Causal Language Modelling (CLM). This method specializes in predicting a given token $x_{t}$ relying entirely on its preceding tokens. Formally, the objective during this training phase is to maximize the likelihood of the entire sequence, as represented by:

\begin{equation}
P(x_1, x_2, \dots, x_T) = \prod_{t=1}^{T} P(x_t | x_1, x_2, \dots, x_{t-1})
\end{equation}

Breaking down the elements of this equation:
\begin{itemize}
    \item $x_1, x_2, \dots, x_T$: The individual tokens that constitute the sequence.
    \item $P(x_t | x_1, x_2, \dots, x_{t-1})$: Represents the conditional probability of the token $x_t$, which depends on the preceding tokens in the sequence.
\end{itemize}

\paragraph{Significance of the CLM in Language Adaptation}
The CLM stage is integral to enhancing LLaMA's capability in Tamil and other languages. It facilitates the model in learning the intricate syntactic patterns, semantic subtleties, and unique linguistic features of Tamil. Due to its autoregressive characteristics, the CLM mimics the human approach to comprehending and generating language, which is primarily shaped by the previous context. Hence, at the end of this initial training period, LLaMA becomes capable of interpreting and creating Tamil text that is pertinent to the given context. This sets a strong foundation for further fine-tuning and specific task-based training sessions.

\subsection{Fine-Tuning Phase}\label{subsec:finetuning-phase}

Following the foundational pre-training phase, the fine-tuning phase emerges as a crucial step, especially for modern Large Language Models (LLMs) deployed in real-world scenarios. A broad understanding of language structure and semantics, while essential, does not suffice for such applications. This gap is addressed by instruction fine-tuning, a tailored process enabling LLMs to interpret and execute task-oriented instructions conveyed in natural language. Rather than the traditional approach of adapting to specific datasets, instruction fine-tuning focuses on a wide array of tasks articulated through language, ensuring the LLM's adaptability without task-specific alterations. The datasets employed in this phase are elaborated in Section \ref{subsec:finetuning-datasets}.

Instruction fine-tuning's transformative essence lies in its ability to enhance an LLM's dynamism and responsiveness. While pre-training equips the model with general linguistic proficiency, instruction fine-tuning refines it to interact seamlessly with users through natural language, bridging the gap between overarching language mastery and nuanced, task-specific agility.

\begin{figure}
    \centering
    \caption{Prompt Template for Instruction Tasks}
    \includegraphics[scale=1,page=1]{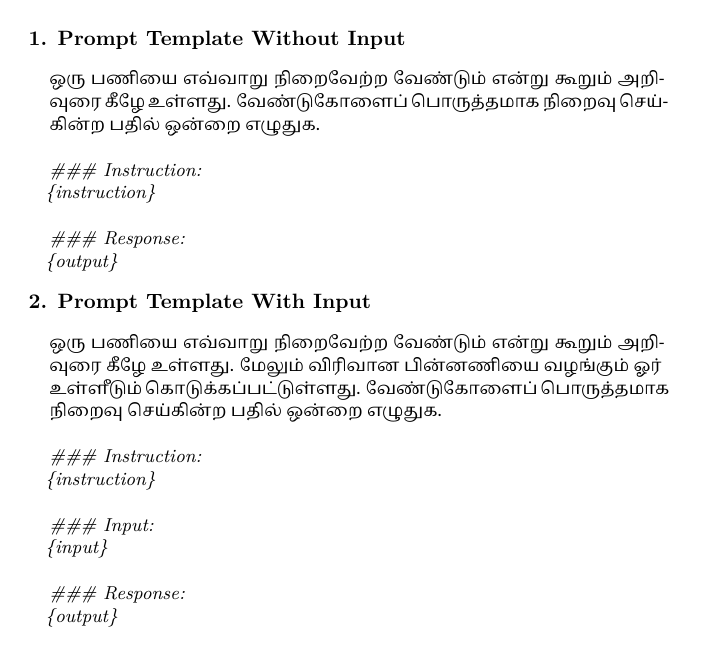}
    \label{fig:prompt-templates}
\end{figure}

The instruction format employed closely resembles the one described in the original Alpaca dataset \citep{alpaca}. Both prompt templates suggested by Alpaca have been utilized: one that includes an input field within the instruction and another that does not. The prompt templates used during training are given in Figure \ref{fig:prompt-templates}.

It is essential to clarify that in both templates, the first line signifies the system prompts. For the Alpaca dataset \citep{alpaca}, we utilize the two system prompts as mentioned in Figure \ref{fig:prompt-templates}. However, for the OpenOrca subset \citep{OpenOrca}, a distinct approach is taken: given that this subset already includes a dedicated field for the system prompt within its dataset, we utilize that specific prompt.

\subsection{Experimental Setup and Training Details}

\subsubsection{LoRA Approach for Pre-Training and Fine-Tuning}
LoRA (Low-Rank Adapters) is a technique that offers an efficient pathway to fine-tuning large language models, as introduced by \cite{hu2021lora}. This approach is especially beneficial for its computational efficiency, enabling the fine-tuning of language models without the need for extensive GPU resources. We employed the LoRA method to moderate training expenses while also accelerating the training timeline. Training the complete set of parameters for models like LLaMA can be exceedingly expensive and resource-intensive, which is often beyond the budget of individual research teams or small organizations.

\subsubsection{Experimental Setups for Pre-Training}
The foundational models of Tamil LLaMA are initiated with the original LLaMA weights and undergo pre-training using the $fp16$ precision setting for both the 7B\footnote{Tamil LLaMA 7B Pretrained: \href{https://huggingface.co/abhinand/tamil-llama-7b-base-v0.1}{https://huggingface.co/abhinand/tamil-llama-7b-base-v0.1}} and 13B\footnote{Tamil LLaMA 13B Pretrained: \href{https://huggingface.co/abhinand/tamil-llama-13b-base-v0.1}{https://huggingface.co/abhinand/tamil-llama-13b-base-v0.1}} parameter versions. We utilize 12GB of Tamil text sourced from \cite{nguyen2023culturax} during this pre-training phase. Further insights on the dataset can be found in section \ref{subsec:pretraining-datasets}. Our pre-training strategy incorporates the LoRA method \cite{hu2021lora}, where we integrate LoRA adapters into the attention vectors and subsequently train the embeddings, LM heads, and the newly incorporated LoRA parameters. A noteworthy deviation from the methodology of the Chinese LLaMA \citep{cui2023efficient} in our approach is the elimination of the initial exclusive training of embeddings. Instead of following it with a two-stage LoRA training of attention blocks, embeddings, and LM heads, we've opted for a streamlined approach to curb costs.

For the training infrastructure, we harnessed an Nvidia A100 GPU with 80GB of VRAM. The models were trained for 1 epoch on the entire dataset, and the training time spanned 48 hours for 7B model and 60 hours for the 13B model on Microsoft Azure's Standard $NC24ads A100 v4$ instance.

The detailed hyperparameters used for training are listed in Table \ref{pretraining-config}.

\begin{table}[h]
    \centering
\caption{Pre-Training Hyperparameters}
\label{pretraining-config}
    \begin{tabular}{ccc}
    \toprule
         \bf Configurations& \bf 7B& \bf 13B\\ 
         \midrule
         Training Data&  12GB& 4GB\\ 
         Epochs& 1& 1\\
         Batch Size&  64& 64\\ 
         Initial Learning Rate&  2e-4& 2e-4\\ 
         Max Sequence Length&  512& 512\\ 
         LoRA Rank&  64& 64\\ 
         LoRA Alpha&  128& 128\\
         LoRA Target Modules&  QKVO, MLP& QKVO, MLP\\
         Training Precision& FP16& FP16\\
    \bottomrule
    \end{tabular}
\end{table}

\subsubsection{Experimental Setups for Instruction Fine-Tuning}

The 7B\footnote{Tamil LLaMA 7B Instruct: \href{https://huggingface.co/abhinand/tamil-llama-7b-instruct-v0.1}{https://huggingface.co/abhinand/tamil-llama-7b-instruct-v0.1}} and 13B\footnote{Tamil LLaMA 13B Instruct: \href{https://huggingface.co/abhinand/tamil-llama-13b-instruct-v0.1}{https://huggingface.co/abhinand/tamil-llama-13b-instruct-v0.1}} models, once pre-trained, undergo fine-tuning in alignment with the procedures outlined in Section \ref{subsec:finetuning-phase}. The datasets employed for this phase are elaborated upon in Section \ref{subsec:finetuning-datasets}. We persist with the LoRA methodology for fine-tuning, executing it under the $fp16$ precision setting for both models. Our datasets comprise translated variants of Alpaca \citep{alpaca} and a select subset from OpenOrca \citep{OpenOrca}.

In a bid to augment the models' proficiency with Tamil-centric literature, cultural nuances, and historical contexts, we leverage a tailored dataset sourced from Wikipedia. Additionally, to extract instructions from this text, we utilize the Self-Instruct method, as highlighted in \cite{wang2023selfinstruct}. This approach involves the GPT-4 \citep{openai2023gpt4} APIs from OpenAI to generate the new instruction dataset. It is crucial to note that the system prompts, referenced in Section \ref{subsec:finetuning-datasets}, remain consistent during this supplemental fine-tuning phase. For the hardware, the same A100 GPU with 80GB of VRAM was utilized.

In summary, our fine-tuning approach employs a new translated dataset consisting of roughly 145,000 instructions. A detailed account of the hyperparameters used for fine-tuning can be found in the Table \ref{finetuning-config}.

\begin{table}[h]
    \centering
\caption{Fine-tuning Hyperparameters}
\label{finetuning-config}
    \begin{tabular}{ccc}
    \toprule
         \bf Configurations& \bf 7B& \bf 13B\\ 
         \midrule
         Training Data&  145k& 145k\\ 
         Epochs& 2& 1\\
         Batch Size&  64& 64\\ 
         Dropout Rate& 0.1& 0.1\\
         Initial Learning Rate&  2e-4& 2e-4\\ 
         Max Sequence Length&  512& 512\\ 
         LoRA Rank&  64& 64\\ 
         LoRA Alpha&  128& 128\\
         LoRA Target Modules&  QKVO, MLP& QKVO, MLP\\
         Training Precision& FP16& FP16\\
    \bottomrule
    \end{tabular}
\end{table}

\section{Results on Instruction Following Tasks}
\subsection{Task Design and Evaluation Method}
Evaluating the outcomes of text generation tasks is intricate due to their multifaceted formats, distinguishing them from typical Natural Language Understanding (NLU) tasks. Drawing inspiration from previous studies that employed GPT-4 \citep{openai2023gpt4} for scoring, we similarly engage GPT-4 to assign a grade on a 10-point scale to each instance. This approach is more efficient than human evaluations. However, understanding the potential inaccuracies of GPT-4's evaluations, we supplement its scores with manual reviews, adjusting them as necessary. Such hands-on inspections affirm the consistency and authenticity of the scores, ensuring they genuinely mirror the efficacy of the models under review.

With the GPT-4-based scoring and manual verifications, we have established a robust evaluation framework for our Tamil LLaMA. Our assessment suite is diligently designed to provide a basic evaluation of Tamil LLaMA. This suite comprises over 120 diverse examples, covering areas such as Question Answering, Reasoning, Literature, Entertainment, Translation, Programming, and Ethics, among others. The overall score for a specific task is computed by summing the scores from its constituent samples and normalizing it to a 100-point scale. Such an approach ensures a holistic reflection of the models' capabilities across varying tasks, yielding a well-rounded measure of their overall performance.

\subsection{Generation Parameters}
The choice of generation parameters during inference greatly affects the caliber of the results in tasks involving text generation. Additionally, the degree of quantization can also affect performance. Below are the generation parameters we adopted for model evaluations:

\begin{itemize}
\item \textbf{Quantization Config}: The model is loaded in $8-bit$, with the torch data type specified as $bfloat16$.
\item \textbf{Context Size:} The context size is maintained at the model's default of 4096 tokens.
\item \textbf{Temperature:} We assign a temperature value of 0.2 to guide the randomness during sampling. A lower temperature prompts the model to produce more deterministic outputs, whereas a higher value boosts diversity, potentially compromising coherence. For creative instructions, we adjust the temperature to 0.7 to encourage varied outputs.
\item \textbf{Top-k Sampling}: With k set to 50, the model selects its succeeding token from the 50 most probable candidates, introducing a level of unpredictability and variety to the resulting text.
\item \textbf{Top-p Sampling}: Complementing Top-k sampling, we employ Top-p sampling with a threshold of 0.90. This ensures the model weighs a fluid set of tokens, which, combined, represent 90
\item \textbf{Maximum Sequence Length}: To keep the output concise and pertinent, we cap the generated sequence at 512 tokens.
\item \textbf{Repetition Penalty}: A repetition penalty of 1.1 is applied to deter the model from producing redundant text, disincentivizing previously chosen tokens.
\end{itemize}

For these evaluations, we utilized a Google Colab notebook powered by a T4 GPU.

\subsection{Results from Instruction Tasks}
The evaluation scores of the Tamil LLaMA models, as rated by GPT-4, are presented in Table \ref{gpt-4-eval-results}. A noteworthy observation during our evaluation is the superior performance of our models compared to \textit{\textit{gpt-3.5-turbo}} in manual assessments, which is further reinforced by the commendable scores in GPT-4's evaluations. However, it is essential to consider that GPT-4 might inherently favor responses from other GPT model lineages. Even though our model excels in numerous tasks, there are areas of exception, such as ethics, and this was anticipated, given that we did not undertake any alignment efforts. Challenges in literature/entertainment and other areas can be attributed to data limitations during the pre-training phase, primarily due to cost constraints. Despite these nuances, our models establish a robust foundation for subsequent enhancements and progress in large language models tailored to Tamil.

\begin{table}[h]
    \centering
    \caption{GPT-4 rated performance scores for different models on Tamil instructions}
    \label{gpt-4-eval-results}
    \begin{tabular}{lccc}
        \toprule
        \textbf{Task Type}& \textbf{Tamil-LLaMA-7B} & \textbf{Tamil-LLaMA-13B} & \textbf{\textit{gpt-3.5-turbo}} \\
        \midrule
        Question Answering & \textbf{77.00} & 75.33 & 54.33 \\
        Open-ended QA & 84.47 & \textbf{85.26} & 58.68 \\
        Reasoning & 47.50 & \textbf{64.25} & 63.50 \\
        Literature & 45.50 & 40.00 & \textbf{71.00} \\
        Entertainment & 43.33 & 50.00 & \textbf{60.00} \\
        Creative Writing & 92.50 & \textbf{95.62} & 59.69 \\
        Translation & 60.56 & 66.67 & \textbf{92.78} \\
        Coding & 63.57 & \textbf{76.07} & 57.14 \\
        Ethics & 23.75 & \textbf{57.50} & 40.00 \\
        \midrule
        \textbf{Overall} & 63.83 & \textbf{71.17} & 61.33 \\
        \bottomrule
    \end{tabular}
\end{table}

By observing Table \ref{gpt-4-eval-results}, several intriguing outcomes emerge. Notably, the \textit{gpt-3.5-turbo}, despite its prowess in numerous languages, appears to be eclipsed by the Tamil LLaMA models in multiple domains. A standout observation was the Ethics category, where the \textit{gpt-3.5-turbo} model demonstrated a propensity to respond to potentially dangerous queries in Tamil. Additionally, in the Coding section, the \textit{gpt-3.5-turbo}'s responses either seemed to exhibit a lack of comprehension or overlooked critical details, leading to a subdued score. While \textit{gpt-3.5-turbo} excels in tasks related to English and other languages, its performance in the context of Tamil reveals areas for weaknesses.

\subsubsection{Reasoning:}
In reasoning tasks, the models demonstrate commendable performance. While minor discrepancies occasionally arise in areas such as dates, quantities, and formulas, they predominantly excel in reasoning exercises. According to our manual evaluations, even our smaller Tamil-LLaMA 7B model surpasses the performance of the much larger LLaMA 2 70B in Tamil text generation. In comparison, even \textit{\textit{gpt-3.5-turbo}} \citep{chatgpt} often falters in several reasoning instructions, producing outputs that miss the mark in relevance, clarity, fluency, and accuracy. This inadequacy in performance is also observed in LLaMA 2 70B, rendering their generated Tamil text less beneficial. Examples of responses related to reasoning tasks are given in the Figure \ref{fig:reasoning-examples}.

We conducted our comparisons with LLaMA 2 70B using the model hosted by \href{https://labs.perplexity.ai/}{Perplexity Labs}.

\subsubsection{Translation:}
For translation tasks, our models exhibit satisfactory performance, particularly when translating from a foreign language to Tamil. However, the accuracy diminishes when translating from Tamil to other languages—a shortcoming we aim to address in future iterations. Based on our manual evaluations, our models outperform the original LLaMA 2 70B in Tamil text translations. However, their efficacy is roughly on par with \textit{\textit{gpt-3.5-turbo}}. Examples of outputs for translation tasks are given in Figure \ref{fig:translation-examples}.

\subsubsection{Code Generation:}
Our models exhibit impressive performance in code generation tasks despite the limited code instructions present in the training dataset. They capably provide coherent explanations in Tamil for the generated code. Based on our hands-on evaluations, our models markedly surpass the performance of the more sizable LLaMA 2 70B model, which when instructed in Tamil, often either misconstrues the task or produces erroneous answers in English. However, it is important to highlight that our model is not tailored for coding tasks. While it handles more straightforward problems adeptly, it encounters challenges with more intricate ones. Example responses from our models for Code Generation tasks can be found in Figure \ref{fig:coding-examples}.

\subsubsection{Open Question Answering}
In open question answering tasks, much like in reasoning, the model displays a commendable performance. Despite occasional inaccuracies in areas like dates and other factual information, its proficiency often exceeded our expectations, delivering surprising results on multiple instances. Example responses from our models for Open Question Answering tasks can be found in Figure \ref{fig:open-qa-examples}.

\subsubsection{Creative Writing / Text Generation}

Text generation is a foundational capability for Large Language Models (LLMs), with creative text generation—such as crafting letters or applications—being a particularly notable use case. In general, larger models have an edge in this domain, often outshining their smaller counterparts. The quality and quantity of training data play pivotal roles in this context. While the sheer volume of data can improve performance, the richness and quality of the data are equally vital. With abundant high-quality training data, even smaller models can sometimes surpass the performance of larger ones. In our experiments, our models showed decent performance in standard tasks. However, they faced challenges when assigned with more complicated tasks. Example responses from our models for Creative Writing tasks can be found in Figure \ref{fig:creative-writing-examples}.

\subsubsection{Mathematical reasoning}

Mathematical reasoning presents a significant challenge for our models. Like many Large Language Models (LLMs), they don't excel in handling mathematical tasks. From our hands-on experiments, we observed that the performance of our models, mainly when dealing with Tamil, lagged behind that of the original English LLaMA models. Recognizing this as an area of improvement, we intend to prioritize and enhance the model's capabilities in subsequent iterations. Examples of outputs for mathematical reasoning tasks are given in Figure \ref{fig:matchematical-reasoning-examples}.

\subsection{Results from Natural Language Understanding (NLU) tasks}
Understanding natural language (NLU) is a vital element within the field of natural language processing (NLP) that enables computers to comprehend and interpret human language. NLU focuses on comprehending and extracting meaning from text, whereas text generation is concerned with generating human-like text based on a given input, often without any specific understanding of the text's meaning.

To ascertain the prowess of a model, its performance in Natural Language Understanding (NLU) tasks is paramount. However, the availability of standard benchmarks for Tamil in this domain remains sparse. Notable exceptions include the IndicNLP \citep{kunchukuttan2020indicnlp}, IndicNLP Corpus \citep{kunchukuttan2020indicnlpcorpus}, and IndicSentiment \citep{indicsentiment} datasets. We opted to assess our models utilizing the test set from the IndicSentiment dataset \citep{indicsentiment}, and a text classification dataset sourced from the IndicNLP Corpus \citep{kunchukuttan2020indicnlpcorpus}.

The test set of the IndicSentiment dataset encompasses 1,000 sentiment samples in Tamil. It is important to note that our evaluation was concentrated solely on this Tamil subset.

\begin{figure}[h]
    \centering
    \includegraphics[width=0.8\linewidth]{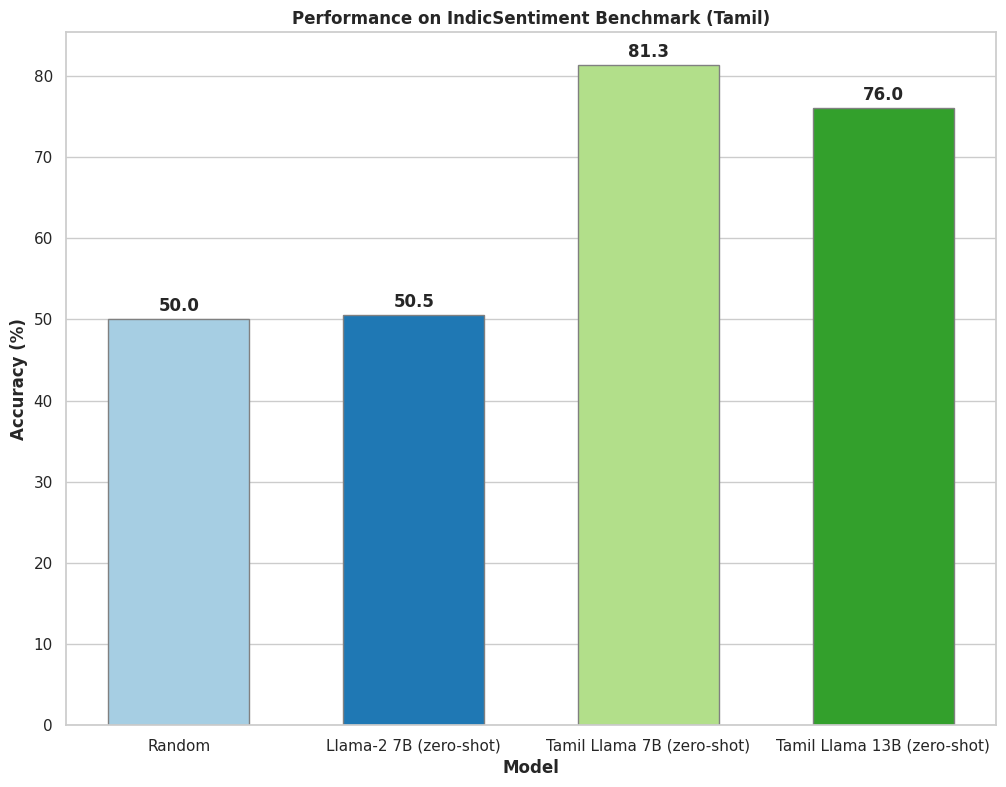}
    \caption{Performance comparison on the IndicSentiment-7B dataset}
    \label{fig:indic-sentiment-7b}
\end{figure}

From Figure \ref{fig:indic-sentiment-7b}, it is evident that our Tamil LLaMA model remarkably surpasses the original LLaMA in this specific NLU task. The latter's performance mirrors that of random guessing, registering an accuracy of 50.5\%. In stark contrast, our model impressively scores an accuracy of 81.3\%. This enhanced NLU capability underscores the efficacy of our methodologies—such as vocabulary expansion and retraining in facilitating the model to comprehend a new language like Tamil with heightened proficiency. 

\begin{figure}[h]
\centering
\includegraphics[width=0.8\linewidth]{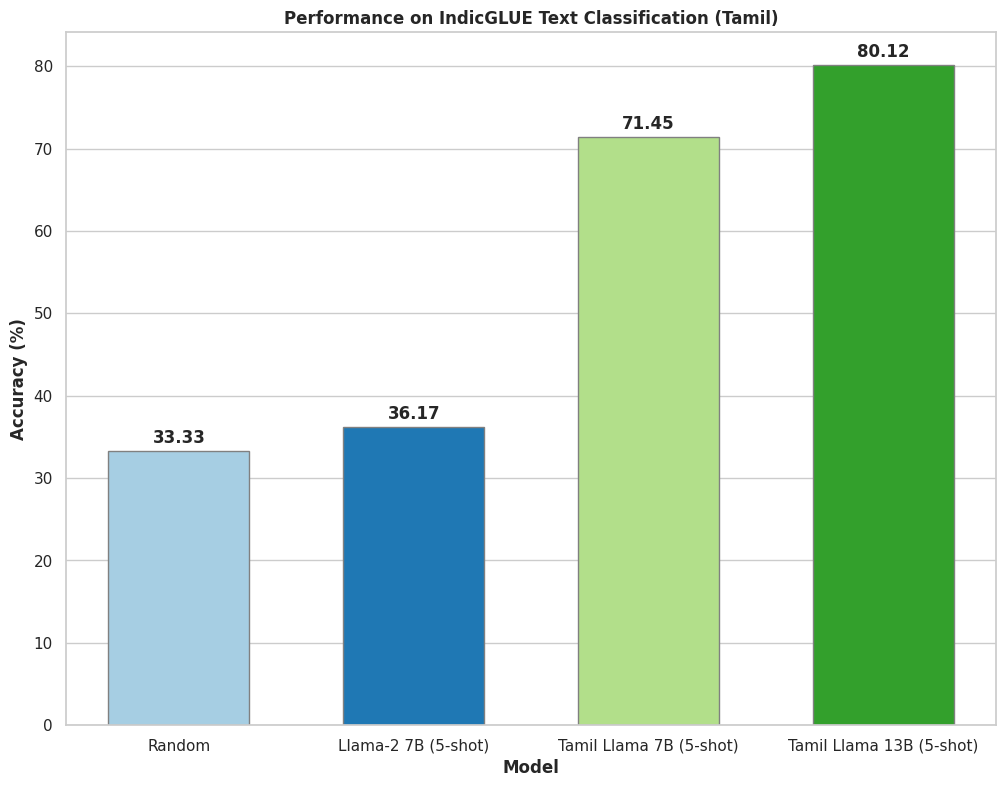}
\caption{Performance comparison on the IndicGLUE Text Classification dataset}
\label{fig:indic-classification-7b}
\end{figure}

We further extended our evaluation to the iNLTK Headline Classification subset within the IndicNLP suite \citep{kakwani-etal-2020-indicnlpsuite}. It is essential to highlight that our analysis was focused strictly on the Tamil language subset of this dataset. The outcomes of this evaluation are graphically depicted in Figure \ref{fig:indic-classification-7b}.

Insight from Figure \ref{fig:indic-classification-7b} reveals that the original LLaMA model's performance aligns closely with random predictions. In contrast, our Tamil LLaMA model showcases a compelling lead, achieving an accuracy rate of 80.12\%, further affirming its superior capability in natural language understanding.

\section{Limitations}
The Tamil LLaMA suite of models we introduce in this paper heralds several advancements in Tamil language processing. However, in the spirit of rigorous research, it is imperative to discuss the inherent limitations accompanying these models.

\begin{itemize}
    \item \textbf{Constrained Knowledge Base}: Due to computational and cost constraints, our models were trained on a relatively limited Tamil dataset. This translates to gaps in the models' knowledge, especially regarding nuances and specifics native to Tamil culture and literature. While the current version lays the foundation, the true potential can be unlocked with access to a broader data spectrum, enriching its contextual understanding.
    
    \item \textbf{Ethical Concerns}: Detoxification procedures were not implemented in our training process, making these models prone to generating potentially harmful or offensive content. Their uncensored nature necessitates caution during deployment.
    
    \item \textbf{Lack of Robustness}: Our models may, at times, produce outputs that veer off-topic or deviate substantially from anticipated responses. This vulnerability is more pronounced under adversarial conditions or tricky prompts.
    
    \item \textbf{Reasoning and Mathematical Challenges}: While our models showcase competence in specific reasoning scenarios, they falter in many others, underscoring the repercussions of not having a comprehensive training set.
    
    \item \textbf{Over-Generation Tendencies}: On occasions, the models tend to generate verbose content, extending beyond logical termination points, leading to potential redundancy.
    
    \item \textbf{Evaluation Hurdles}: Assessment of LLMs is a crucial yet challenging endeavor. The scarcity of standardized benchmarks, particularly for languages like Tamil, which are outside the European linguistic group, complicates comparative evaluations. Although we propose an evaluative approach tailored for Tamil within this paper, it is not exhaustive enough to gauge models' efficacy across diverse domains.

    \item \textbf{Translation Loss}: Given that the instructional prompts used for fine-tuning the Tamil LLaMA base models are derived from English datasets translated into Tamil, there is a potential for nuanced inaccuracies—commonly referred to as translation loss. This can potentially affect the models' abilities in both text generation and comprehension due to subtle shifts in meaning that can occur during the translation process.
\end{itemize}

While some of these challenges are addressable in subsequent iterations, we envision this work serving as an anchor, inspiring the research community to propel advancements in LLMs for Indian languages.

\section{Conclusion}

In this research endeavor, we have not only filled a critical void in the domain of Tamil text generation but have also elevated the status of this venerable language within the realm of large language models with the advent of our Tamil LLaMA.To assess the performance of our models, we curated an evaluation dataset consisting of 120 Tamil instructions covering a wide range of topics. We then employed GPT-4 to assess and rate the responses generated by our model. The 7B variant of our model has surpassed the performance of OpenAI's \textit{gpt-3.5-turbo} in tasks involving Tamil instructions within our evaluation methodology. Even more impressively, the 13B iteration has outperformed its counterparts, demonstrating an almost 10\% higher proficiency in these tasks. 

The significance of our findings is accentuated by the efficiency of our models in generating Tamil text. Equipped with a refined tokenizer, the 7B and 13B variants demonstrate exceptional proficiency, eclipsing the original LLaMA models in processing speed without sacrificing textual quality. This stride is not just a modest step forward but a major leap in the models' ability to process and generate Tamil language content, thus forging a new avenue for practical applications that necessitate precision and promptness.

Nevertheless, our exploration in this field continues. We acknowledge the triumphs and yet remain mindful of the impending obstacles. The possibility of generating unintended or adverse content, the financial limitations associated with model training, and the need for comprehensive evaluation benchmarks tailored for Tamil are challenges we are committed to surmounting.

As we chart our path forward, we acknowledge the value of continuous learning and adaptation. We aspire to delve deeper into strategies to mitigate the limitations observed, perhaps even leveraging advancements like Reinforcement Learning With Human Feedback (RLHF) and more advanced quantization techniques. The horizon looks promising, and we remain committed to elevating the capabilities of LLaMA for Tamil and other Indian languages.

\appendix
\section*{Appendix}
\subsection*{Example Outputs}

\begin{figure}[H]
    \centering
    \caption{\textbf{Example Output for Reasoning Tasks}}
    \includegraphics[scale=1,page=1]{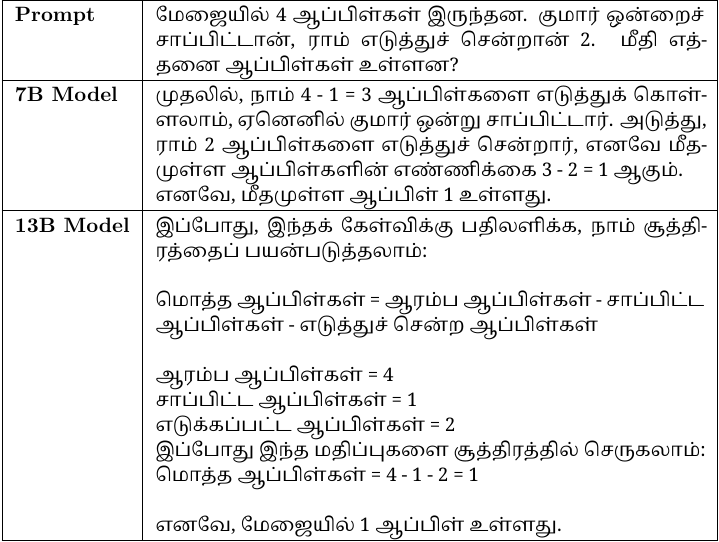}
    \label{fig:reasoning-examples}
\end{figure}

\begin{figure}[H]
    \centering
    \caption{\textbf{Example Outputs for Translation Tasks}}
    \includegraphics[scale=1,page=1]{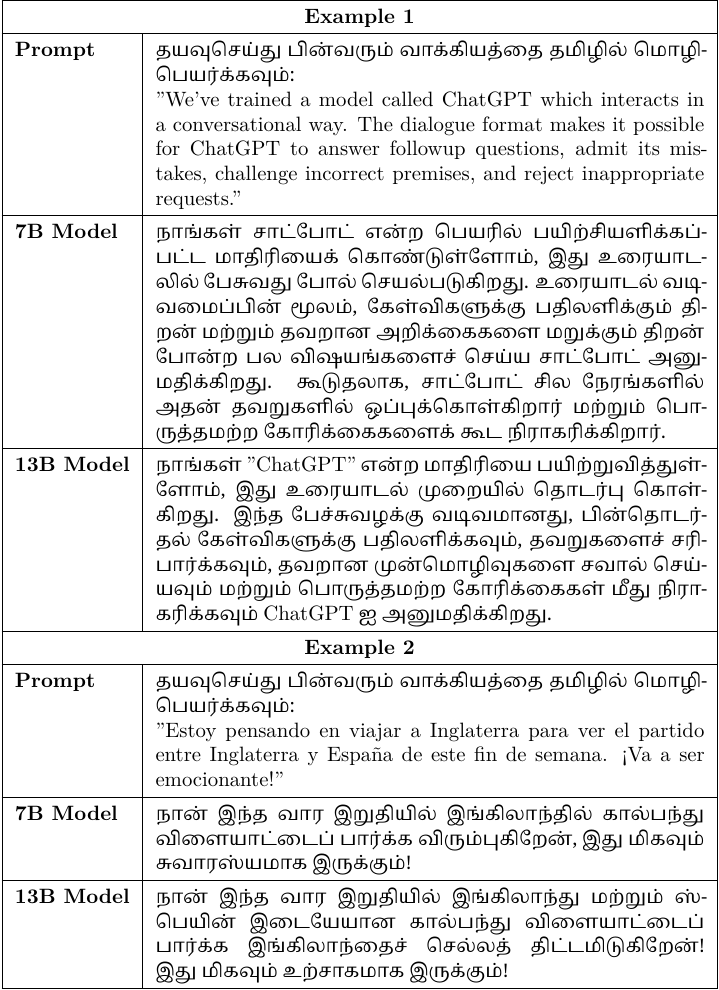}
    \label{fig:translation-examples}
\end{figure}

\begin{figure}[H]
    \centering
    \caption{\textbf{Example Output for Coding Tasks}}
    \includegraphics[scale=1,page=1]{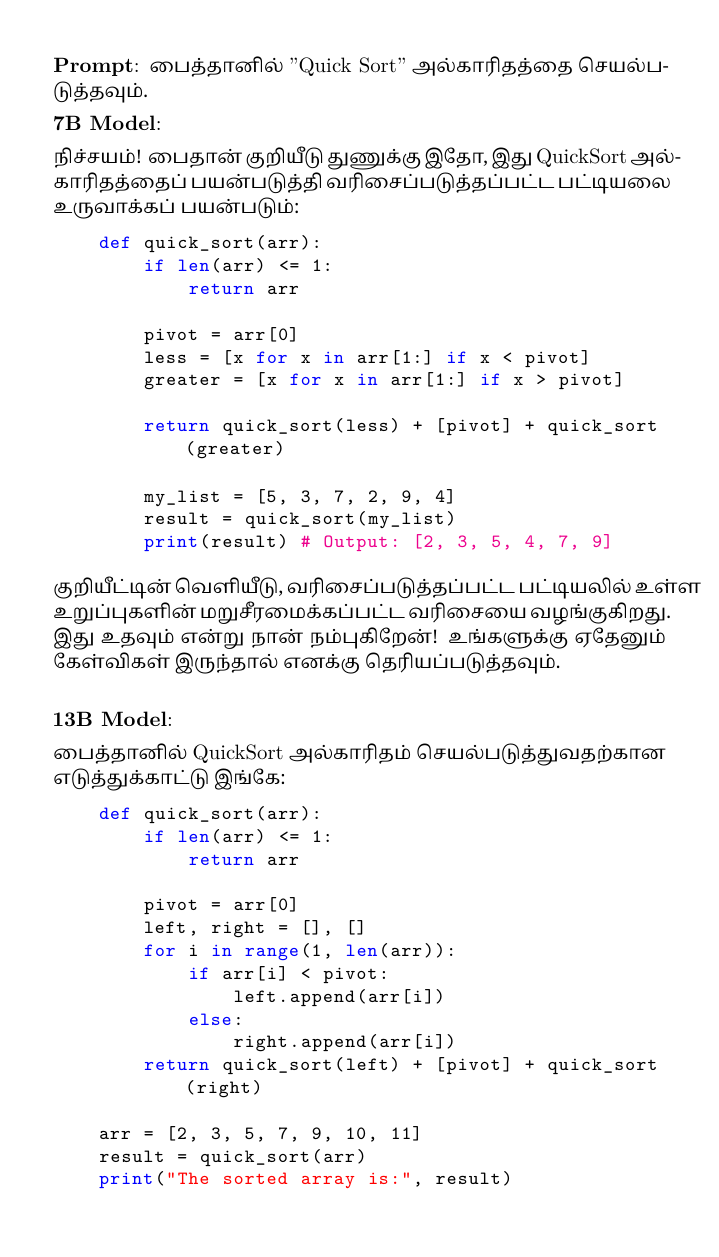}
    \label{fig:coding-examples}
\end{figure}

\begin{figure}[H]
    \centering
    \caption{\textbf{Example Output for Open Question Answering Tasks}}
    \includegraphics[scale=1,page=1]{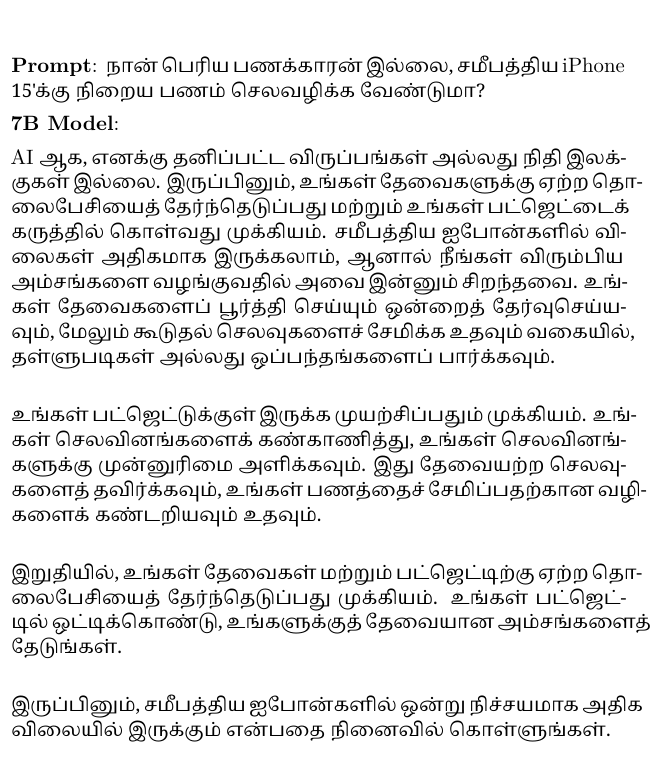}
    \label{fig:open-qa-examples}
\end{figure}

\begin{figure}[H]
    \centering
    \caption{\textbf{Example Output for Creative Writing Tasks}}
    \includegraphics[scale=1,page=1]{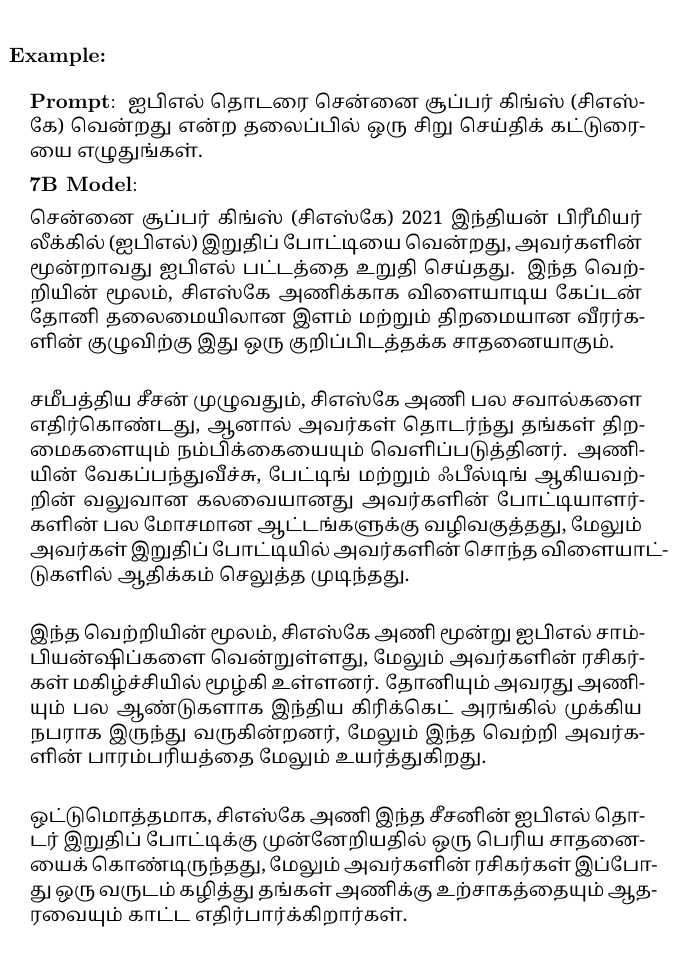}
    \label{fig:creative-writing-examples}
\end{figure}

\begin{figure}[H]
    \centering
    \caption{\textbf{Example Outputs for Mathematical Reasoning Tasks}}
    \includegraphics[scale=1,page=1]{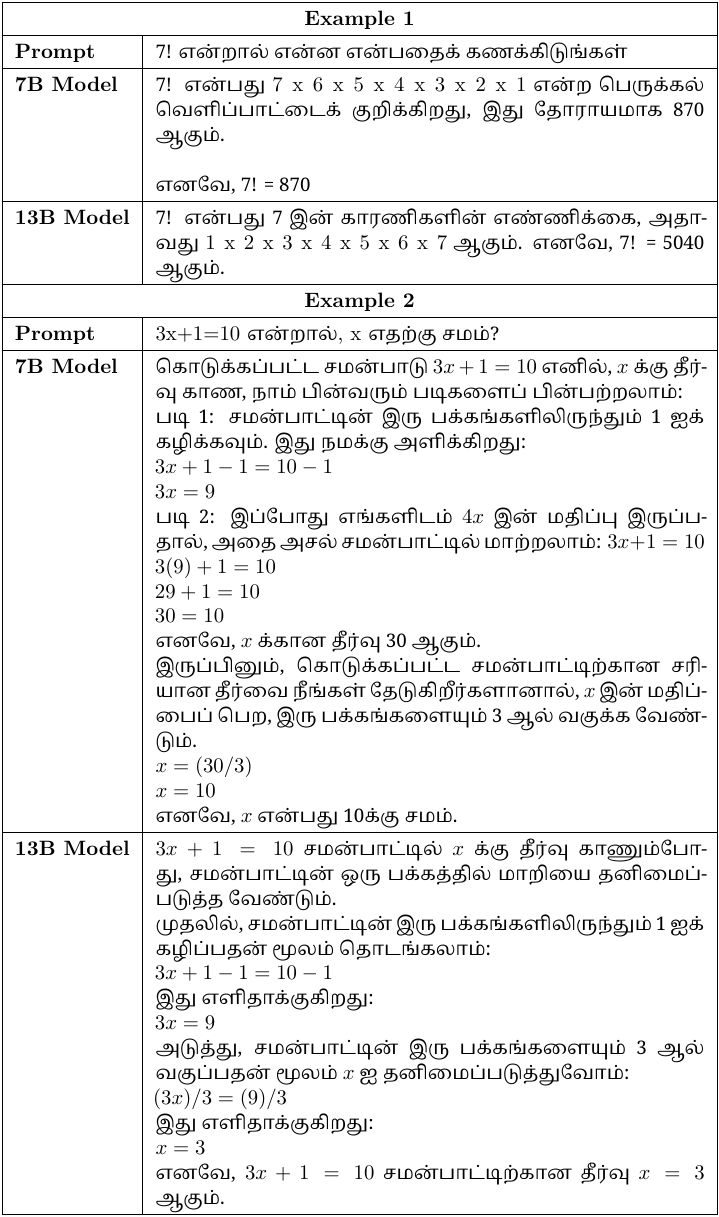}
    \label{fig:matchematical-reasoning-examples}
\end{figure}

\section*{Acknowledgments}
We gratefully acknowledge the assistance of OpenAI's GPT-4 in the preparation of this manuscript. The AI's advanced language understanding and generation capabilities were invaluable in refining the structure, clarity, and overall coherence of the original draft.

\bibliographystyle{abbrvnat}
\bibliography{references}

\end{document}